\documentclass[10pt,twocolumn,letterpaper]{article}

\usepackage{iccv}
\usepackage{times}
\usepackage{epsfig}
\usepackage{graphicx}
\usepackage{amsmath}
\usepackage{amssymb}

\usepackage[ruled]{algorithm2e}
\usepackage{appendix}
\usepackage{booktabs}
\usepackage{dsfont}
\usepackage{enumitem}
\usepackage{subcaption}

\usepackage[pagebackref=true,breaklinks=true,letterpaper=true,colorlinks,bookmarks=false]{hyperref}

\iccvfinalcopy 

\def\iccvPaperID{1529} 

\ificcvfinal\fi

\newcommand{\figref}[1]{Figure~\ref{#1}}
\newcommand{\tabref}[1]{Table~\ref{#1}}

\newcommand{\algref}[1]{Alg.~\ref{#1}}
\newcommand{\equref}[1]{Equ.~\ref{#1}}

\begin{document}

\title{Take More Positives: An Empirical Study of Contrastive Learing\\ in Unsupervised Person Re-Identification}

\author{Xuanyu He$^1$, Wei Zhang$^1$, Ran Song$^1$, Qian Zhang$^1$, Xiangyuan Lan$^2$, Lin Ma$^3$\\
$^1$Shandong University \
$^2$Hong Kong Baptist University \
$^3$Meituan \\
}

\maketitle
\ificcvfinal\thispagestyle{empty}\fi

\begin{abstract}
Unsupervised person re-identification (re-ID) aims at closing the performance gap to supervised methods. These methods build reliable relationship between data points while learning representations. However, we empirically show that the reason why they are successful is not only their label generation mechanisms, but also their unexplored designs. By studying two unsupervised person re-ID methods in a cross-method way, we point out a hard negative problem is handled implicitly by their designs of data augmentations and PK sampler respectively. In this paper, we find another simple solution for the problem, \ie, taking more positives during training, by which we generate pseudo-labels and update models in an iterative manner. Based on our findings, we propose a contrastive learning method without a memory back for unsupervised person re-ID. Our method works well on benchmark datasets and outperforms the state-of-the-art methods. Code will be made available.
\end{abstract}

\section{Introduction}

Unsupervised person re-identification (re-ID) aims at learning discriminative identity-level representations from unlabeled person images.
In common person re-ID settings, manual annotations are provided as supervisory signals to reduce the distance between instances of the same person (\ie, positive pairs) and increase the distance between instances of different persons (\ie, negative pairs).
Since instances of an identity vary significantly under different camera views, it is challenging to model the distribution of identities in the fully unsupervised case.

When ground truths are unavailable, we require a label generating mechanism to discover the relationship between instances for learning identity-level representations.
In the literature, there are several methods for generating pseudo-labels.
For example, by maintaining a memory bank of the entire dataset, MMCL~\cite{wang2020unsupervised} mines reliable neighbors of samples and then generates multi-labels for training.
Another general way of generating pseudo-labels is clustering~\cite{ge2020self,lin2019bottom,zeng2020hierarchical}.
Incorporating different strategies (\eg, self-paced~\cite{ge2020self} and hierarchical~\cite{zeng2020hierarchical} strategies) into clustering, unsupervised person re-ID is closing the performance gap to supervised methods.

The core idea of these existing methods is to build reliable relationships with others for each data point in the dataset while learning representations.
With the help of reliable pseudo-labels, memory-based methods (\eg, MMCL~\cite{wang2020unsupervised} and SpCL~\cite{ge2020self}) are able to learn identity-level representations in a unsupervised contrastive learning manner.
We believe that their good performance rely on their label generation mechanisms a lot, by which uncertain examples are handled properly during training.

However, we empirically show that an important problem behind their success has been handled implicitly but not explored.
Since their learning processes are approximate to self-supervised learning at the start of training, distances between positive examples inevitably increase and the difficulty of discovering true positive examples increases as well.
In other words, true positive examples become hard to be discovered as positives in the unsupervised cases.
Different from these hard examples caused by illuminations, deformations, occlusions and other intra-class variations, which might be well known as hard negatives, this \emph{hard positive problem} discussed in this paper is caused by the act of learning process itself when ground truths are unavailable.
Therefore, if this \emph{hard positive problem} is not deal with by specific designs, the representation space will develop to an uniform feature distribution~\cite{wang2020understanding} and models degenerate to instance-level representation learning.

\begin{figure*}[!t]
\centering
\includegraphics[width=0.75\linewidth]{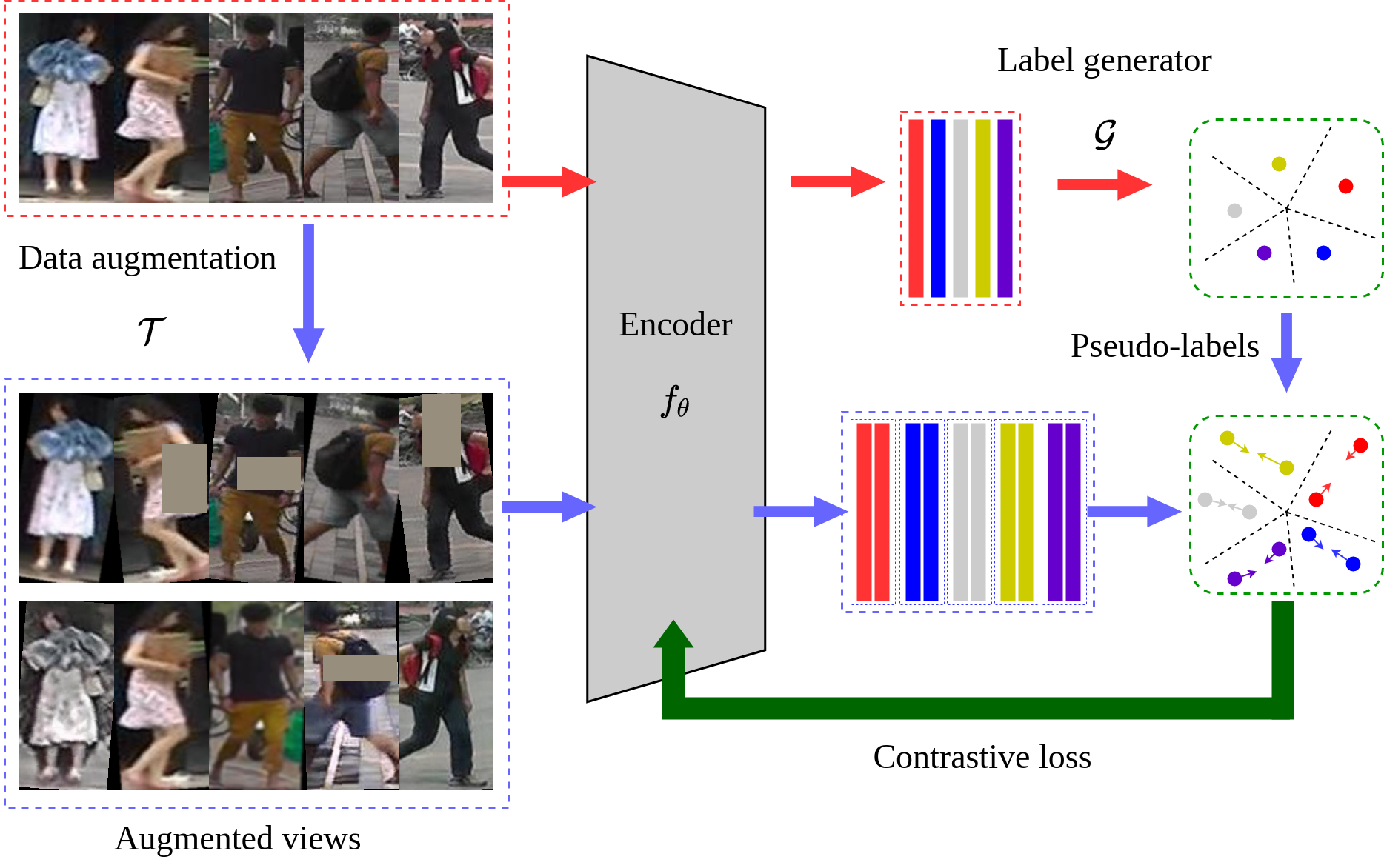}
\caption{Our simple contrastive learning method for unsupervised person re-ID.
A data augmentation module is applied to each example to obtain two views augmented in different manners from the same image.
The encoder network $f_\theta$ is used to extract features from images.
Our method explores the relationship between data points with pseudo-labels $y$ generated by $\mathcal{G}$.}
\label{fig:method}
\end{figure*}

In this paper, we report that there is another simple solution can solve the hard positive problem besides existing designs. 
Based on our findings, our model can directly learn identity-level representations in a contrastive manner without a memory bank.
We illustrate this simple method in \figref{fig:method}.
Our method outperforms the state-of-the-art methods in field of unsupervised person re-ID.
The main contributions of this work are summarised as follows:
1) We empirically study two successful unsupervised person re-ID methods.
Our study shows that an important \emph{hard negative problem} has been handled by their designs but not explored;
2) We empirically show that the reason why these unsupervised person re-ID methods are successful is not only their reliable label generation mechanisms but also they prevent models from overfitting to an uniform feature distribution;
3) We propose a simple contrastive learning method for unsupervised person re-ID.
Our experimental results show that our solution for the hard negative problem can produce meaningful results.
Based on our findings, our method outperforms the state-of-the-art methods.




\section{Related work}

Person re-ID is a widely studied computer vision problem.
Supervised learning is highly successful in person re-ID.
Advanced techniques~\cite{yao2019deep,luo2019bag,li2020multi}, \eg, attention mechanism~\cite{li2018harmonious,zhang2019multi,chen2019abd,zhang2020relation}, perform well on person re-ID datasets.
Meanwhile, the community also has an interest in unsupervised learning methods for person re-ID.

\subsection{Unsupervised domain adaptation}

Efforts have been made to develop unsupervised domain adaptation (UDA) methods for person re-ID, which transfer the learned knowledge from the labeled source domain to the unlabeled target domain~\cite{wei2018person,deng2018image,zhong2018generalizing,zhong2019invariance,yu2019unsupervised,fan2018unsupervised,fu2019self,ge2020mutual,ge2020self,zhao2020unsupervised}.
Generative methods, such as PTGAN~\cite{wei2018person} and SPGAN~\cite{deng2018image}, transfer person images from source domain to target domain, and then use transferred images for training.
Some works, such as MAR~\cite{yu2019unsupervised}, MMT~\cite{ge2020mutual}, and SpCL~\cite{ge2020self}, utilize the source dataset as a reference to generate pseudo-labels, supervising the training of models.
Different from these UDA works, our method uses no labeled sample yet achieves good results.

\subsection{Unsupervised person re-identification}

There are some works focusing on unsupervised person re-ID without any labeled examples from source domain~\cite{lin2019bottom,lin2020unsupervised,wang2020unsupervised,zeng2020hierarchical}.
BUC~\cite{lin2019bottom} and HTC~\cite{zeng2020hierarchical} generated pseudo-labels by hierarchical clustering methods, and updated models with the pseudo-labels using classification loss or triplet loss.
To avoid the effects of noisy pseudo-labels, MMCL~\cite{wang2020unsupervised} formulated unsupervised person re-ID as a multi-label classification problem, by maintaining a memory bank of all the instances in the dataset.
Similar to MMCL, SSL~\cite{lin2020unsupervised} proposed to treat unsupervised person re-ID as a softened classification task.
Our work differs with most previous works as our method works in a contrastive learning manner. 
In the literature, MMCL~\cite{wang2020unsupervised} and unsupervised version of SpCL~\cite{ge2020self} are also contrastive learning methods.
Generally, we differ from them that our method does not require a memory bank of the entire dataset.

\subsection{Contrastive learning}

Contrastive learning methods~\cite{he2020momentum,chen2020simple,grill2020bootstrap} have been successful in self-supervised representation learning.
These approaches learn instance-level discriminative representation by contrasting positive pair against negative pair without supervisory signals.
Our work share many similarities with these state-of-the-art contrastive learning methods, in particular using views differently augmented from an image as positive pairs.
Different from self-supervised representation learning that uses only a single positive, our contrastive learning method achieves good results in the field of person re-ID by taking more positive examples.

In the field of person re-ID, SpCL~\cite{ge2020self} also adopted a contrastive learning loss with a running hybrid memory.
Meanwhile, MMCL~\cite{wang2020unsupervised} can be treated as a variant of contrastive learning from the view of objective loss.
Despite their success in unsupervised person re-ID, there is an important \emph{hard positive problem} caused by contrastive learning remaining unexplored.
In this paper, we empirically show that they handled this problem implicitly by different choices.



\section{Empirical study}

In this section, we empirically study two contrastive learning methods for unsupervised person re-ID on Market-1501~\cite{zheng2015scalable} dataset: MMCL~\cite{wang2020unsupervised}~\footnote{MMCL can be treated as a contrastive learning method since it has a similar loss formulation with multi-labels.} and SpCL$^\dagger$~\cite{ge2020self}~\footnote{We denote the unsupervised version of SpCL as SpCL$^\dagger$.}.
Performance is evaluated by cumulative matching characteristic (CMC) curve and mean average precision (mAP)~\cite{zheng2015scalable}.
We pay attention to what might contribute to promising performance in the filed of unsupervised person re-ID.

\subsection{Data augmentation}

\begin{table}[!t]
\centering
\resizebox{\linewidth}{!}{ %
\begin{tabular}{@{}lcccc@{}}
\toprule
method & augmentations & PK sampler & mAP & Rank-1 \\
\midrule
MMCL & $\mathcal{T}_{\text{MMCL}}$ &  &
49.9 & 81.4 \\
SpCL$^\dagger$ & $\mathcal{T}_{\text{SpCL$^\dagger$}}$ & \checkmark &
72.4 & 88.7 \\
\midrule
MMCL & $\mathcal{T}_{\text{SpCL$^\dagger$}}$ & &
6.2  & 19.0 \\
SpCL$^\dagger$ & $\mathcal{T}_{\text{MMCL}}$ & \checkmark &
65.7 & 86.3 \\
\midrule
SpCL$^\dagger$ & $\mathcal{T}_{\text{SpCL$^\dagger$}}$ & &
13.9 & 30.3 \\
SpCL$^\dagger$ & $\mathcal{T}_{\text{MMCL}}$ & & 
33.4 & 58.6 \\
\bottomrule
\end{tabular}
}
\caption{Effect of data augmentations and PK sampler.
For reproducing each method, we follow the hyper-parameter and augmentation recipes in original paper.
}
\label{tab:aug_pk}
\end{table}

In \tabref{tab:aug_pk} we study the effect of data augmentations.
Unsupervised person re-ID methods are successful using different data augmentation recipes.
However, the behaviors of data augmentations have not been studied in a cross-method way.
We denote their data augmentation modules as $\mathcal{T}_{\text{MMCL}}$ and $\mathcal{T}_{\text{SpCL$^\dagger$}}$ respectively.
Generally, $\mathcal{T}_{\text{SpCL$^\dagger$}}$ is a standard and simple person re-ID augmentation module while $\mathcal{T}_{\text{MMCL}}$ is a more sophisticated one.
Detailed augmentations are described in the supplement.

We note that performance of MMCL drops significantly if using data augmentations from SpCL$^\dagger$.
This observation is unsurprising since MMCL acts as an instance-level representation learning method (\ie, contrastive learning in a self-supervised manner) at the first several epochs, spreading instances of the same identity apart.
In this case, a simple data augmentation module such as $\mathcal{T}_{\text{SpCL$^\dagger$}}$ makes the model quickly overfit to a uniform feature distribution and the true positive examples lose the probability to be discovered as positives when it starts to mine neighbors.
In contrast, various augments expand the real data distribution and slow down this process.

However, SpCL$^\dagger$ is also approximate to instance-level learning at the starting, since data points are unable to be well clustered that time.
It finally archives good performance with its simple data augmentations.
We believe that SpCL$^\dagger$ reduces the negative impact of contrastive learning in unsupervised cases by an important design, \ie, PK sampler.

\subsection{PK sampler}

PK sampler is a common setting in the field of supervised person re-ID~\cite{hermans2017defense,luo2019bag}.
It randomly chooses $P$ identities and $K$ instances for each identity from the dataset as a batch during training.
Meanwhile, this sampling strategy is also widely used in the unsupervised cases~\cite{zeng2020hierarchical,ge2020self}.
When ground truths are unavailable, PK sampler is implemented on top of pseudo-labels.

In \tabref{tab:aug_pk} we remove the PK sampler in SpCL$^\dagger$.
Without the sampling strategy, the performance of SpCL$^\dagger$ drops a lot, similar to the behavior of MMCL when its $\mathcal{T}_{\text{MMCL}}$ is replaced.
Our explanation is that SpCL$^\dagger$ can reform data distribution and focus on the organized batches at every iteration by the PK sampler.
With this design, models are offered with certain positives and negatives for each data point and escape from the force that pushes the point away from other uncertain examples.

\subsection{Summary}

We have empirically shown the reason why these unsupervised person re-ID methods are successful.
In unsupervised scenarios, person re-ID models have to deal with a force that spreads potential positives apart during training, especially at the start.
This hard positive problem is caused by the act of learning in the unsupervised scenario, because many instances are treated as individual classes.
The two methods properly handled the negative impact of that force and prevented models from overfitting to an uniform feature distribution~\cite{wang2020understanding} at the start of training by a \emph{sophisticated data augmentation module} or a \emph{specific sampling strategy}.

In this work, we find that there is another solution can produce meaningful results by avoiding that problem: when we search positive samples in the representation space for each data point, just take more positives against the inevitable force.







\section{Method}

In this paper, we propose a contrastive learning method, namely Take More Positives (TMP), for unsupervised person re-ID.

As illustrated in~\figref{fig:method}, our architecture takes as input two randomly augmented views $x_i$ and $x_j$ from an image $x$, produced by a data augmentation module $\mathcal{T}$.
The two views are processed by an encoder network $f_\theta$.
Denoting the two output vectors as $z_i = f_\theta(x_i)$ and $z_j = f_\theta(x_j)$, we optimize the encoder $f_\theta$ by a contrastive learning loss with the pseudo-label $y$ generated by $\mathcal{G}$ as follows:
\begin{equation}
\label{eq:loss_i}
\mathcal{L}_i = \sum_{j=1}^{2N} \mathds{1}_{[j \neq i \land y_j = y_i]} \mathcal{L}_{ij} 
\end{equation}
\begin{equation}
\label{eq:loss_ij}
\mathcal{L}_{ij} = - \log \frac{\exp (s_{i, j} / \tau)}
{\sum_{k=1}^{2N} \mathds{1}_{[k \neq i \land (k = j \lor y_k \neq y_i)]} \exp (s_{i, k} / \tau)}
\end{equation}
\begin{equation}
\label{eq:sim}
s_{i, j} = \frac{z_i \cdot z_j}{{\Vert z_i \Vert}_2 {\Vert z_j \Vert}_2}
\text{,}
\end{equation}
where ${\Vert \cdot \Vert}_2$ is $\ell_2$-norm.
The final total loss is averaged over all image views.
\algref{alg} summaries our method.

\begin{algorithm}[!t]

\SetKwInOut{Input}{input}
\SetKwInOut{Return}{return}

\Input{batch size $N$, data augmentation module $\mathcal{T}$, encoder network $f_\theta$ and pseudo-label generator $\mathcal{G}$.}

\For{sampled batch $\{x_k\}_{k = 1}^{N}$}{
    \For{$k \in \{i, \dots, N\}$}{
    generate pseudo-label $y_k = \mathcal{G} (f_\theta (x_k))$ \\
    }

    \For{$k \in \{1, \dots, N\}$}{
    the first augmented view \\
    $t \sim \mathcal{T}$ \\
    $\tilde{x}_{2k - 1} = t (x_k)$ \\
    $z_{2k - 1} = f_\theta (\tilde{x}_{2k - 1})$ \\
    $y_{2k - 1} = y_k$ \\
    the second augmented view \\
    $t^\prime \sim \mathcal{T}$ \\
    $\tilde{x}_{2k} = t^\prime (x_k) $ \\
    $z_{2k} = f_\theta (\tilde{x}_{2k})$ \\
    $y_{2k} = y_k$
    }

    \For{$i \in \{1, \dots, 2N\}$ and $j \in \{1, \dots, 2N\}$}{
    compute $s_{i, j}$ by \equref{eq:sim}
    }
    
    \For{$i \in \{1, \dots, 2N\}$}{
    compute $\mathcal{L}_i$ by \equref{eq:loss_ij} and \equref{eq:loss_i}
    }
    
    $\mathcal{L} = \frac{1}{2N} \sum_{i = 1}^{2N} \mathcal{L}_i$ \\
    update encoder $f_\theta$ to minimize $\mathcal{L}$
}

\Return{encoder network $f_\theta$}

\caption{Learning algorithm}
\label{alg}

\end{algorithm}

If $\mathcal{G}$ outputs the index of an instance in the dataset as its label (\ie, each instance is an individual class), our architecture degenerates to self-supervised learning framework SimCLR~\cite{chen2020simple}.
When we use ground truths during training, our architecture is similar to supervised contrastive learning (SCL)~\cite{khosla2020supervised}.
Note that our architecture is not identical to SCL with pseudo-labels.
Formally, SCL calculates $\mathcal{L}^{sup}_{ij}$ as follows:
\begin{equation}
\label{eq:loss_ij_scl}
\mathcal{L}^{sup}_{ij} = - \log \frac{\exp (s_{i, j} / \tau)}
{\sum_{k=1}^{2N} \mathds{1}_{[k \neq i ]} \exp (s_{i, k} / \tau)}
\text{.}
\end{equation}
For a positive pair $\{x_i, x_j\}$ (\equref{eq:loss_ij} \emph{vs.} \equref{eq:loss_ij_scl}), SCL loss calculates all positive pairs while we exclude others except $\{x_i, x_j\}$.
As a result, SCL loss function will push other positives (\eg, $x_p$) away from $x_i$ when the model is optimized for $\{x_i, x_j\}$.
Though it will pull $x_p$ back when the loss term turns to $\mathcal{L}^{sup}_{ip}$, it leads to inefficient optimization.

\paragraph{Baseline settings.}
We use a modified ResNet-50 as the default encoder $f_\theta$ as in~\cite{luo2019bag,ge2020self,wang2020unsupervised}.
Implementation details are described in the supplement.
We use the following settings for unsupervised training unless specified:

\begin{itemize}[leftmargin=*]
\item Label generator $\mathcal{G}$. We generate pseudo-labels $y$ for instances in the dataset by a label generator $\mathcal{G}$, \ie, $y \triangleq \mathcal{G}(f_\theta(x))$.
Therefore, views augmented in different manners from the same image, $\tilde{x}_i$ and $\tilde{x}_j$, share the same pseudo-label $y$.
We adopt DBSCAN~\cite{ester1996density,schubert2017dbscan} with Jaccard distance~\cite{zhong2017re} as our label generator $\mathcal{G}$.
Generally, the clustering process is controlled by the maximum distance between neighbors $\epsilon$.
The unclustered instances are treated as individual classes.
Other implementation details of clustering are in the supplement.

\item Optimizer. We use SGD for training models.
We use a learning rate of $lr \times \text{BatchSize} / 256$, with a base $lr = 0.1$.
The batch size is 256 by default.
The learning rate has a cosine decay schedule.
The weight decay is 0.0001 and the SGD momentum is 0.9.
We train models for 100 epochs in ablation experiments unless specified.
\end{itemize}

\section{Experiments}

In this section, we report experimental results on Market-1501~\cite{zheng2015scalable} dataset.
For evaluation, we resize images to $256 \times 128$ and normalize them with RGB mean and standard deviation.
We evaluate person re-ID performance on $\ell_2$ normalized features by euclidean distance.
No postprocessing (\eg, re-ranking~\cite{zhong2017re}) is used.

\paragraph{Baseline.}
Our baseline uses data augmentations from SpCL$^\dagger$ during training.
No PK sampler is applied in the baseline.
The clustering threshold $\epsilon$ in $\mathcal{G}$ is 0.6 following~\cite{ge2020self}.
Our baseline has a 8.1\% mAP and 21.9\% Rank-1 accuracy on the test set (\tabref{tab:tmp}).

\begin{table}[!t]
\centering
\resizebox{\linewidth}{!}{
\begin{tabular}{@{}cccccc@{}}
\toprule
& $\epsilon$ & $\mathcal{T}$ & PK sampler &  mAP & Rank-1 \\
\midrule
baseline & 0.6 & $\mathcal{T}_{\text{SpCL$^\dagger$}}$ & &
8.1 & 21.9 \\
\textbf{a} & 0.75 & $\mathcal{T}_{\text{SpCL$^\dagger$}}$ & &
41.0 & 63.6 \\
\textbf{b} & 0.75 &  $\mathcal{T}_{\text{SpCL$^\dagger$}}$ & \checkmark &
49.1 & 71.4 \\
\textbf{c} & 0.75 & $\mathcal{T}_{\text{MMCL}}$ & \checkmark &
66.7 & 85.9 \\
\textbf{d}& 0.75 & $\mathcal{T}_{\text{MMCL}}$ & &
53.4 & 74.9 \\
\bottomrule    
\end{tabular}
}
\caption{Experimental results of our method on Market-1501~\cite{zheng2015scalable}.
}
\label{tab:tmp}
\end{table}

\paragraph{Take more positives.}
We simply consider more positives for each data point by increasing $\epsilon$ in $\mathcal{G}$.
The maximum distance between neighbors $\epsilon$ controls the clustering process.
With a larger $\epsilon$, data points will treat more examples as their positives and then more instances will be clustered.

In \tabref{tab:tmp}, we first increase $\epsilon$ to 0.75.
The results show that taking more positives indeed improves performance.
We also find that performance can be further improved by PK sampler and data augmentations.
Although this process (\ie, take more positives) introduces noisy examples, our results suggest that contrastive learning is able to fix this problem during training.

\begin{figure*}[!t]
\centering
\includegraphics[width=0.75\linewidth]{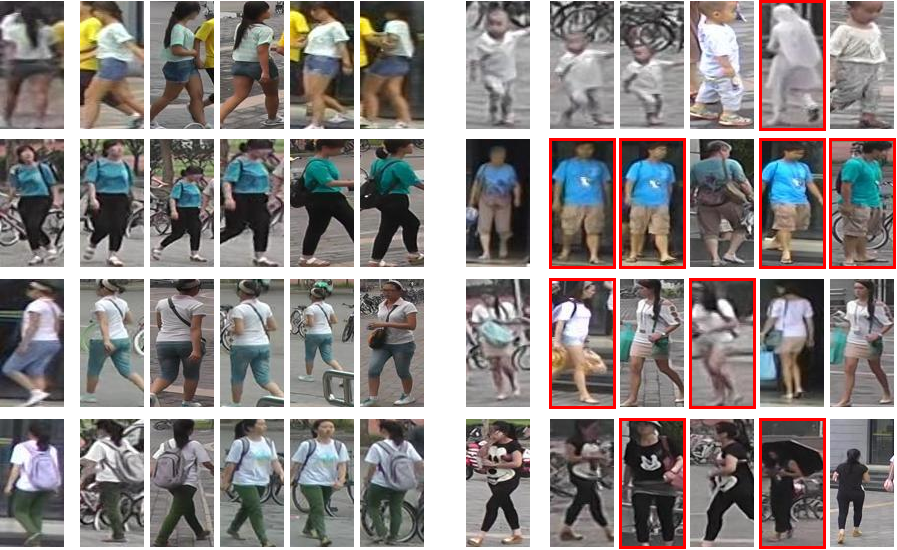}
\caption{Ranking results on Market-1501 test set.
Red boxes denote the wrong ranking results.
We present good ranking results as well as bad ranking results here.
By visualizing ranking results, we find that these wrong re-ID results have similar appearances with queries.
}
\label{fig:ranking}
\end{figure*}

In the following, we further improve our TMP performance based on other designs.
In \figref{fig:ranking} we visualize the final person re-ID results of our method.

\paragraph{Data augmentation.}
We describe data augmentation using PyTorch notations.
We use the sophisticated data augmentation module $\mathcal{T}_{\text{MMCL}}$ as the start.
In \tabref{tab:augs}, we first remove the color augmentation \texttt{ColorJitter} from $\mathcal{T}_{\text{MMCL}}$ and find that color augmentation is not helpful when CamStyle is applied.
Meanwhile, \texttt{RandomGrayscale} with an applying probability of 0.2 leads to a significant drop on performance.
The result suggests that appearance is an important element in unsupervised person re-ID.
We add blurring augmentation \texttt{GaussianBlur} with an applying probability of 0.5, which has a Gaussian kernel with std in $[0.1, 2.0]$.
In final, it comes to our data augmentation module $\mathcal{T}_{\text{TMP}}$ after our trials.

We further report lesion studies on composition of data augmentation operations $\mathcal{T}_{\text{TMP}}$.
To better understand the effects of individual data augmentations in unsupervised person re-ID, we evaluate the performance by removing or adding data augmentations individually in \tabref{tab:augs}.
The details of our implementation of data augmentations are in the supplement.
Note that \texttt{RandomErasing}~\cite{zhong2020random}, which works well in supervised person re-ID, is still important for achieving good performance in the unsupervised scenario.

\begin{table}[!t]
\centering
\begin{tabular}{@{}lcc@{}}
\toprule
$\mathcal{T}$ & mAP & Rank-1 \\
\midrule
$\mathcal{T}_1: \mathcal{T}_{\text{MMCL}}$ & 66.7 & 85.9 \\
$\mathcal{T}_2: \mathcal{T}_{\text{MMCL}} - \texttt{ColorJitter}$ &
67.5 & 86.6 \\
$\mathcal{T}_3: \mathcal{T}_2 + \texttt{RandomGrayscale}$ &
21.0 & 45.8 \\
$\mathcal{T}_4: \mathcal{T}_2 + \texttt{GassuianBlur}$ &
68.3 & 86.8 \\
\midrule
$\mathcal{T}_{\text{TMP}}$ & 68.3 & 86.8 \\
\ - \texttt{RandomHorizontalFlip} & 64.9 & 83.3 \\
\ - \texttt{RandomRotation} & 67.3 & 85.4 \\
\ - \texttt{RandomErasing} & 59.8 & 82.5 \\
\bottomrule    
\end{tabular}
\caption{Data augmentation experiments on Market-1501~\cite{zheng2015scalable}.
}
\label{tab:augs}
\end{table}

\paragraph{Temperature.}
Temperature $\tau$ in \equref{eq:loss_ij} effectively weights different examples, and an appropriate temperature can help the model learn from hard negatives.
In \tabref{tab:tau}, we test different temperature values $\tau$ in the proposed contrastive loss.
Results show that the performance is significantly worse without proper temperature scaling.
Moreover, a very small temperature $\tau = 0.01$ fails to converge.
In this work, we use a fixed temperature $\tau = 0.05$ for all the experiments.

\begin{table}[t!]
\centering
\begin{tabular}{ccc}
\toprule
$\tau$ & mAP & Rank-1 \\
\midrule
0.01 & 0.1  & 0.1  \\
0.05 & 68.3 & 86.8 \\
0.1  & 65.6 & 84.9 \\
0.2  & 39.9 & 65.1 \\
0.5  & 13.7 & 34.4 \\
\bottomrule
\end{tabular}
\caption{Results of different choices of temperature $\tau$ on Market-1501.
}
\label{tab:tau}
\end{table}

\paragraph{Clustering.}
After improving the performance of our TMP based on other designs, we explore the behaviors of our TMP under different clustering thresholds again.
In our experiments, we find that a loose clustering leads to better performance as shown in~\tabref{tab:epsilon}, consistent to our findings.
We also report the number of final clusters on the training set.
Note that there are 751 person identities in the Market-1501 training set, which is very close to our results.

\begin{table}[t!]
\centering
\begin{tabular}{cccc}
\toprule
$\epsilon$ & mAP & Rank-1 & clusters \\
\midrule
0.5   & 55.2 & 84.5 & 1373 \\
0.6   & 56.2 & 85.2 & 1377 \\
0.65  & 56.5 & 85.6 & 1195 \\
0.7   & 68.1 & 87.4 & 843  \\ 
0.75  & 68.3 & 86.8 & 734  \\
\bottomrule
\end{tabular}
\caption{Results of different clustering threshold $\epsilon$ on Market-1501.
}
\label{tab:epsilon}
\end{table}

\begin{figure*}[t!]
\centering
\begin{subfigure}{0.33\linewidth}
\includegraphics[width=\linewidth]{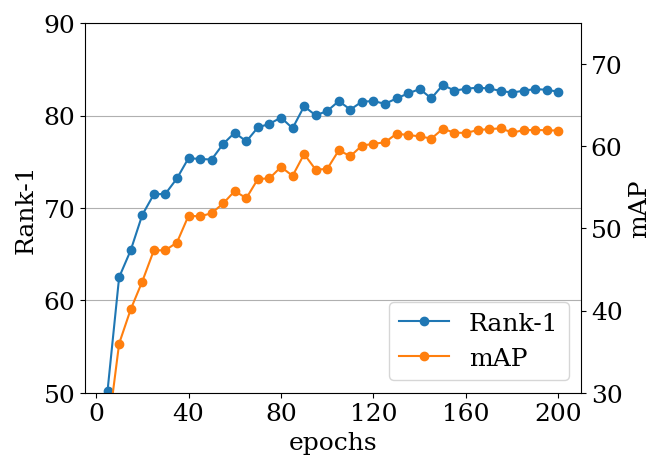}
\caption{$N = 64$}
\label{fig:training.n64}
\end{subfigure}  
\hfill
\begin{subfigure}{0.33\linewidth}
\includegraphics[width=\linewidth]{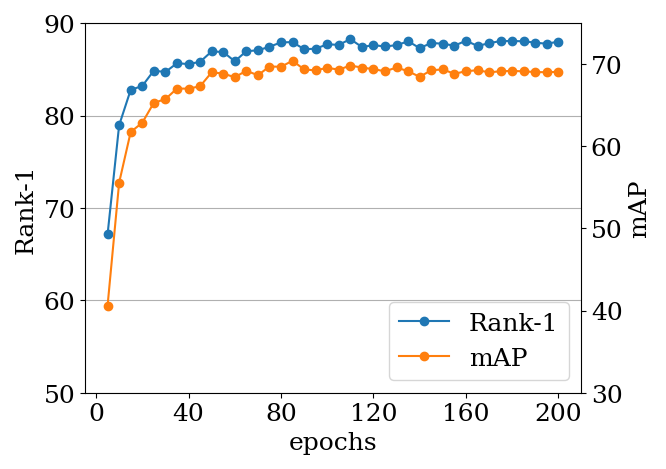}
\caption{$N = 256$}
\label{fig:training.n256}
\end{subfigure} 
\hfill
\begin{subfigure}{0.33\linewidth}
\includegraphics[width=\linewidth]{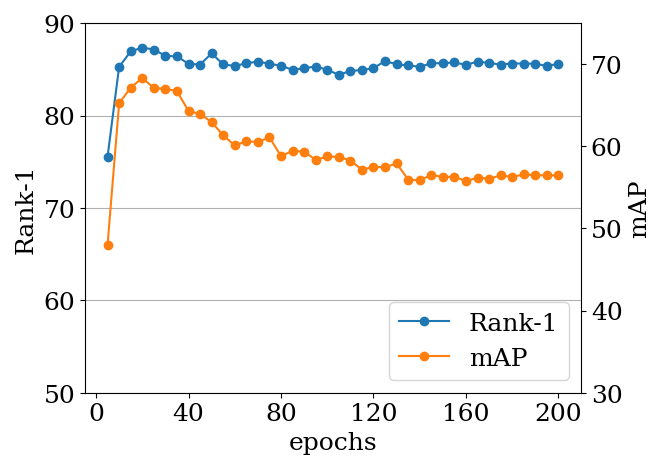}
\caption{$N = 1024$}
\label{fig:training.n1024}
\end{subfigure}
\caption{Experimental results of batch sizes on Market-1501.
Models are trained for 200 epochs.
}
\label{fig:training}
\end{figure*}

\paragraph{Batch size.}
In \figref{fig:training} we show the impact of batch size.
Larger batch sizes, which allow us to train models with more positives and negatives, have a significant advantage over small ones ($N = 64$ \emph{vs.} $N = 256$).
However, we find that a very large batch size like 1024 instead deteriorates person re-ID performance.
We believe that a number of possible positive examples are treated as negatives with such a large batch size from the start of training, pushing instances of the same identity away from each other and overwhelming models.
After distances between positives is beyond the clustering threshold, they finally lose the possibility to be clustered as a common class during training.
We also note that mAP should be a more important index to evaluate performance of the unsupervised person re-ID.

\paragraph{Multi-epoch label generation.}
In \tabref{tab:interval} we present re-ID performance of our method under different interval of updating clustering-based labels.
We show that our method has a stable performance even if we update pseudo-labels every 4 epochs.
Considering the good clustering performance by taking more positives (\tabref{tab:epsilon}), this experiment suggests that the quality of clustering has been good at the early time of training.

\begin{table}[!t]
\centering
\begin{tabular}{ccc}
\toprule
$E$   & mAP   & Rank-1  \\
\midrule
1     & 68.3  & 86.8    \\
2     & 68.1  & 87.0    \\
4     & 67.1  & 86.5    \\
\bottomrule
\end{tabular}
\caption{Experimental results of label generation interval $E$ on Market-1501.}
\label{tab:interval}
\end{table}

\paragraph{Extra data.}
In \tabref{tab:extra_data} we use extra data that share the same source with Market-1501 for unsupervised training~\cite{zheng2015scalable}.
Unlike transfer learning settings, extra data from the same source does not bring a domain gap problem.
There are 500k images in total.
For a comparison, Market-1501 has about 12k images for training.
We train our model for 100 epochs.
PK sampler is applied by default.
We use two data augmentation modules, \ie, $\mathcal{T}_{\text{SpCL$^\dagger$}}$ and $\mathcal{T}_{\text{TMP}^-}$ (we remove CamStyle from $\mathcal{T}_{\text{TMP}}$).

\tabref{tab:extra_data} shows that a large amount of images, even distractors, can improve the re-ID performance in the unsupervised case.
Moreover, with such a amount of images, the effect from data augmentations is not significant.

\begin{table}[!t]
\centering
\begin{tabular}{@{}lccc@{}}
\toprule
$\mathcal{T}$ & extra data & mAP & Rank-1 \\
\midrule
$\mathcal{T}_{\text{SpCL$^\dagger$}}$ & & 49.1 & 71.4 \\
$\mathcal{T}_{\text{SpCL$^\dagger$}}$ & \checkmark & 69.1 & 87.3 \\
$\mathcal{T}_{\text{TMP}^-}$ & \checkmark & 73.4 & 88.9 \\
\bottomrule    
\end{tabular}
\caption{Experimental results of using extra data.
}
\label{tab:extra_data}
\end{table}

\paragraph{Label generator alternation.}
We replace the clustering algorithm in label generator $\mathcal{G}$ with PUL method~\cite{fan2018unsupervised}.
PUL clusters examples by k-means and only selects reliable data instances for training.
In this experiment, we follow their clustering implementation ($K = 750$ for k-means) but treat these unreliable examples as distinct classes.
By this way, we are able to train the total dataset with our contrastive learning loss.
As in~\cite{fan2018unsupervised}, we use cosine similarity to compute the distance between examples.
Let $\lambda$ denote the reliability threshold.
Therefore, a larger $\lambda$ means a stricter sample selection.

Results are reported in~\tabref{tab:label}. 
Our framework is still effective when we use other clustering algorithms for label generation.
Moreover, it is also observed that a loose clustering leads to better performance than others in our framework.
As discussed above, contrastive learning loss pushes other instances away from the start of training, \ie, their cosine similarities decrease during training.
With a higher $\lambda$, examples from the same identity will have no chance to be clustered together after a number of iterations.

\begin{table}[!t]
\centering
\begin{tabular}{@{}ccc@{}}
\toprule
$\lambda$ & mAP  & Rank-1 \\
\midrule
0.55  & 68.5 & 86.7 \\
0.6   & 68.3 & 86.3 \\
0.65  & 59.0 & 85.1 \\ 
0.7   & 55.9 & 81.4 \\
0.75  & 43.8 & 73.8 \\
\bottomrule
\end{tabular}
\caption{Results of different choices of $\lambda$ on Market-1501.}
\label{tab:label}
\end{table}
\section{Comparisons}

In this section, we compare the TMP with the state-of-the-art methods on benchmark datasets.
Beyond person re-ID results, we also compare the methodologies of some related works.

\subsection{Main results}

\begin{table*}[!t]
\centering
\begin{tabular}{lcccccc}
\toprule
      &            &             &
\multicolumn{2}{c}{Market-1501~\cite{zheng2015scalable}} & \multicolumn{2}{c}{DukeMTMC-reID~\cite{ristani2016performance}} \\
method & batch size & memory bank & mAP & Rank-1 & mAP & Rank-1 \\
\midrule
BUC~\cite{lin2019bottom} & 16 & &  
38.3 & 66.2 & 27.5 & 47.4 \\
HTC~\cite{zeng2020hierarchical} & 64  & &
56.4 & 80.0 & 50.7 & 69.6 \\
SSL~\cite{lin2020unsupervised} & 16 & & 
37.8 & 71.7 & 28.6 & 52.5 \\
MMCL~\cite{wang2020unsupervised} &
128 & $\checkmark$ &
45.5 & 80.3 & 40.2 & 65.2 \\
SpCL$^\dagger$~\cite{ge2020self} &
64 & $\checkmark$ &
73.1 & 88.1 & - & - \\
\midrule
TMP (100 epochs) & 256 & &
68.3 & 86.8 & 53.2 & 73.2 \\
TMP (200 epochs) & 256 & &
74.1 & 89.5 & 58.3 & 72.8 \\
\bottomrule
\end{tabular}
\caption{Comparisons with the state-of-the-art unsupervised person re-ID methods on Market-1501~\cite{zheng2015scalable} and DukeMTMC-reID~\cite{ristani2016performance}.
}
\label{tab:main}
\end{table*}

We compare the TMP with the state-of-the-art unsupervised person re-ID methods including BUC~\cite{lin2019bottom}, HTC~\cite{zeng2020hierarchical}, SSL~\cite{lin2020unsupervised}, MMCL~\cite{wang2020unsupervised}, and SpCL$^\dagger$~\cite{ge2020self} in \tabref{tab:main} on Market-1501~\cite{zheng2015scalable} and DukeMTMC-reID~\cite{ristani2016performance} datasets.
We report the best results for training 200 epochs. 

\tabref{tab:main} shows that the results and the main properties of these methods.
TMP is trained with a batch size of 256 without a memory bank of the entire dataset.
It has the best performance among all methods.

\subsection{Methodology comparisons}

\paragraph{Relation to SCL.}
Our TMP is conceptually analogous to ``SCL~\cite{khosla2020supervised} with pseudo-labels''.
We have explained the main difference between \equref{eq:loss_ij} and \equref{eq:loss_ij_scl} formally.
Both methods share a similar idea: take many positives for each point in addition to many negatives using labels.
However, our method considers more positive examples by loosing the requirement of clustering, even though it introduces noisy examples.

The results of our SCL reproduction in the unsupervised cases on Market-1501 is in \tabref{tab:scl_simclr}.
As we explained before, SCL loss does not exclude other positive pairs when computing $\mathcal{L}_{ij}$, resulting in inefficient optimization.

\begin{table}[!t]
\centering
\begin{tabular}{@{}lccc@{}}
\toprule
& pseudo-labels & mAP & Rank-1 \\
\midrule
SCL    & \checkmark & 62.0 & 81.4 \\
SimCLR &            & 11.4 & 30.9 \\
TMP    & \checkmark & 68.3 & 86.8 \\
\bottomrule
\end{tabular}
\caption{Comparison with SCL~\cite{khosla2020supervised} and SimCLR~\cite{chen2020simple} on Market-1501~\cite{zheng2015scalable}.}
\label{tab:scl_simclr}
\end{table}

\begin{figure}[!t]
\centering
\begin{subfigure}{0.45\linewidth}
\includegraphics[width=\linewidth]{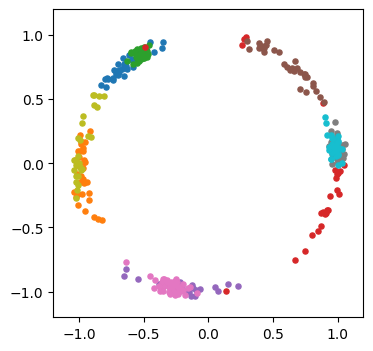}
\end{subfigure}
\begin{subfigure}{0.45\linewidth}
\includegraphics[width=\linewidth]{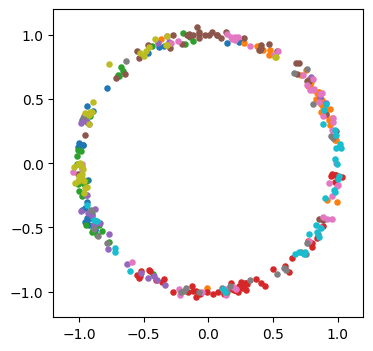}
\end{subfigure}
\caption{Visualization on representations learned by our TMP (left) \emph{vs.} SimCLR using only one positive (right).
Data points can not be clustered in the representation space using one positive for person re-ID tasks.}
\label{fig:simclr}
\end{figure}

\paragraph{Relation to SimCLR.}
SimCLR~\cite{chen2020simple} is a contrastive learning framework designed for self-supervised representation learning.
SimCLR can be treated as our counterpart which uses only a single positive during training.
The the comparison on our SimCLR reproduction for person re-ID is in \tabref{tab:scl_simclr}.
\figref{fig:simclr} shows the comparison on the representations learned from dataset.

\paragraph{Relation to SpCL$^\dagger$.}
SpCL$^\dagger$~\cite{ge2020self} relies on a memory bank of the entire dataset for optimization.
Conceptually, TMP can be thought of as SpCL$^\dagger$ without memory bank and self-paced clustering.
In this paper, we have empirically shown the reason why SpCL$^\dagger$ is successful in the field of unsupervised person re-ID.
Moreover, we propose another solution to take more positives during training, which in fact share a similar motivation with SpCL$^\dagger$.

\section{Conclusion}

Unsupervised learning is closing the performance gap to supervised methods in many fields of computer vision.
In this paper, we empirically study existing contrastive learning methods for unsupervised person re-ID.
We empirically point out a hard positive problem caused by the force from contrastive learning, which spreads positives apart from the start of training.
Successful methods deal with this problem by different designs.
However, the reason why these designs work well has not been explored before.

Beyond the empirical study, we propose another solution to reduce the negative impact of contrastive learning in unsupervised cases.
The strength of our method suggests that, despite the noisy pseudo-labels used during training, taking more positive examples against that force is still beneficial to mime the relationship of person images in this scenario.
We hope that our work could inspire further research on unsupervised person re-ID and other unsupervised computer vision tasks.

{\small
\bibliographystyle{ieee_fullname}
\bibliography{references.bib}
}

\newpage

\appendix

\section{Implementation details}

\paragraph{Encoder.}
We use a modified ResNet-50~\cite{he2016deep} as the default encoder in this work following~\cite{luo2019bag,wang2020unsupervised,ge2020self}.
The last stride of the downsampling operation in the original network is set to 1 (original ResNet-50: 2).
The classifier (a fully-connected layer) is removed.
We add a batch normalization (BN) layer~\cite{ioffe2015batch} at the end of ResNet-50 as in~\cite{luo2019bag,wang2020unsupervised,ge2020self}.
The output is 2048-d.

\paragraph{Data augmentation.}
We describe data augmentations using PyTorch notations.

\begin{itemize}[leftmargin=*]
\item $\mathcal{T}_{\text{MMCL}}$:
First, CamStyle~\cite{zhong2018camera} is applied to the original image if its camid is different from a random chosen one. 
Geometric augmentation is \texttt{RandomResizedCrop} with scale in [0.64, 1.0], \texttt{RandomHorizontalFlip} and \texttt{RandomRotation} with a degree in [-10, 10].
Color augmentation is \texttt{ColorJitter} with \{brightness, contrast, saturation\} strength of \{0.2, 0.2, 0.2\}.
After normalizing image with mean and standard deviation, \texttt{RandomErasing} is applied with scale in [0.02, 0.33] and a probability of 0.5.

\item $\mathcal{T}_{\text{TMP}}$:
We fix the probability of applying CamStyle~\cite{zhong2018camera} to 0.5, which is different from $\mathcal{T}_{\text{MMCL}}$~\cite{wang2020unsupervised}.
Geometric augmentation is as same as $\mathcal{T}_{\text{MMCL}}$.
$\mathcal{T}_{\text{TMP}}$ uses no color augmentation.
We find that color augmentation may conflict with CamStyle in unsupervised cases.
Without CamStyle, \texttt{ColorJitter} can bring a slight improvement on performance.
\texttt{RandomErasing} is applied in final as same as $\mathcal{T}_{\text{MMCL}}$.

\item $\mathcal{T}_{\text{SpCL$^\dagger$}}$:
Geometric augmentation is \texttt{Resize} of a given size (256, 128), \texttt{RandomHorizontalFlip}, and \texttt{RandomCrop} with padding of 10 and the target size (256, 128).
Finally, \texttt{RandomErasing} is applied as the other two.
\end{itemize}

\paragraph{PK sampler.}
PK sampler randomly choose $P$ identities and $K$ instances for each identity from the dataset.
We set $K = 4$ in all the experiments as in~\cite{luo2019bag,ge2020self}.
The $P$ is then decided by the batch size, \ie, $P = \text{BatchSize} / K$.
In the unsupervised cases, we implement PK sampler following~\cite{ge2020self},
\ie, adopting $K = 1$ when the instance is an individual class (unclustered sample).

\paragraph{Clustering.}
We use DBSCAN~\cite{ester1996density,schubert2017dbscan} with Jaccard distance~\cite{zhong2017re} for clustering.
The implementation of DBSCAN is based on \texttt{scikit-learn}.
The minimal number of samples in a neighborhood for a point to be considered as a core point it set to 4.
For calculating Jaccard distance, we set $k1$ to 30 and $k2$ to 6~\cite{zhong2017re}.


\end{document}